\documentclass[runningheads]{llncs}
\usepackage{graphicx}

\usepackage{amsmath, array, float}

\usepackage{tikz}
\usetikzlibrary{positioning,arrows,calc,backgrounds,pgfplots.groupplots}
\usepackage{pgfplots}
\pgfplotsset{compat=newest}
\usepackage{xcolor}

\usepackage{multirow}
\usepackage{booktabs}
\usepackage[flushleft]{threeparttable}
\newcolumntype{C}[1]{>{\centering\arraybackslash }p{#1}}

\usepackage{makecell}
\usepackage{marvosym}

\usepackage[square, comma, numbers]{natbib}
\makeatletter \renewcommand\@biblabel[1]{#1.} \makeatother
\usepackage[hyperfootnotes=false,breaklinks=true,hidelinks,colorlinks=true,citecolor=blue,urlcolor=blue,linkcolor=blue]{hyperref}

\def\@fnsymbol#1{\ensuremath{\ifcase#1\or *\or \dagger\or \ddagger\or
		\mathsection\or \mathparagraph\or \|\or **\or \dagger\dagger
		\or \ddagger\ddagger \else\@ctrerr\fi}}

\newcommand{\ssymbol}[1]{^{\@fnsymbol{#1}}}
\usepackage{etoolbox}
\newcommand{\repthanks}[1]{\textsuperscript{\ref{#1}}}
\makeatletter
\patchcmd{\maketitle}
{\def\thanks}
{\let\repthanks\repthanksunskip\def\thanks}
{}{}
\patchcmd{\@maketitle}
{\def\thanks}
{\let\repthanks\@gobble\def\thanks}
{}{}
\newcommand\repthanksunskip[1]{\unskip{}}
\makeatother

\begin{document}
\title{RP-DQN: An application of Q-Learning to Vehicle Routing Problems}
\titlerunning{}
%
\author{Ahmad Bdeir\thanks{Equal contribution\protect\label{X}} \and
	Simon Boeder\repthanks{X} \and
	Tim Dernedde$^{\text{(\Letter)}}$\repthanks{X} \and
	Kirill Tkachuk\repthanks{X} \and
	Jonas K. Falkner \and Lars Schmidt-Thieme}
\authorrunning{ }
%
\institute{University of Hildesheim, 31141 Hildesheim, Germany\\
	\email{\{bdeir,boeders,dernedde,tkachuk\}@uni-hildesheim.de}\\
	\email{\{falkner,schmidt-thieme\}@ismll.uni-hildesheim.de}}
\maketitle              
\begin{abstract}
	In this paper we present a new approach to tackle complex routing problems with an improved state representation that utilizes the model complexity better than previous methods. We enable this by training from temporal differences. Specifically Q-Learning is employed. We show that our approach achieves state-of-the-art performance for autoregressive policies that sequentially insert nodes to construct solutions on the CVRP. Additionally, we are the first to tackle the MDVRP with machine learning methods and demonstrate that this problem type greatly benefits from our approach over other ML methods.
	
\end{abstract}
\section{Introduction}
Routing problems are very important in business and industry applications. Finding the best routes for delivery vehicles, finding the best pick up order of trading goods in a warehouse or the optimal machine paths in a manufacturing factory are just a few examples for such problems. Due to their importance, many of these problems have been thoroughly studied and the traditional operations research community has identified a wide variety of problem types including various constraints and developed many heuristics for solving these \citep{r1,r2,r3,r5,r6,r7,r8}. Recently, the Machine Learning (ML) community has proposed to learn heuristics with models instead of handcrafting them. The main advantage of ML methods is that once initially trained, they can solve new problem instances very quickly, while traditional heuristics solve every problem individually which in the general case takes up significantly more time.

As optimal labeled solutions are expensive or intractable to compute, usually Reinforcement Learning (RL) is employed to optimize these models. Specifically, various forms of the REINFORCE \citep{williams} algorithm were applied \citep{bello, nazari, kool, falkner}. A recent method by \citet{kool} has proven to find good solutions on a variety of problem types including the capacitated vehicle routing problem (CVRP). Their architecture consists of an encoder and decoder setup that constructs the tour sequentially. The encoder computes a representation of the graph and its nodes and is run once in the beginning of the tour construction. Then the decoder is applied on the embeddings repeatedly to add nodes to the solution until the tour is finished. This method is limited to static node features, since the encoder is applied only once. Using dynamic features, and hence a more rich representation, would require the encoder to be run at every step. This quickly becomes memory-expensive when using REINFORCE, as all intermediate representations have to be stored over the whole episode in order to execute backpropagation. 

Motivated by this limitation, we propose a model that enables the use of dynamic features in the encoder. Encoder and decoder are then applied in every step. To make this memory-feasible, we optimize our model with a temporal-difference algorithm. When learning only on single transitions, the intermediate representations do not have to be stored for the whole episode. Specifically, we will use Q-Learning \citep{dqn}.

We show that the improved state representation leads to performance improvements on the CVRP. Additionally, we show that our approach is more sample efficient which can be an important factor when environments become more complex and expensive to compute. Finally, we show, when extending to more complicated problems like the Multiple Depot Vehicle Routing Problem (MDVRP), the improved state representation is of even greater importance. As to our knowledge we are the first to explore the MDVRP with RL-methods, thus we extended the model by \citet{kool} to solve the MDVRP and compare to our approach and several established OR-methods.

\section{Related work}
The operations research community has studied a plenitude of different routing problems over the last 70 years. Various different soft and hard constraints, objective functions and other problem properties like stochasticity have been considered and many optimal and heuristic solution approaches with and without formal guarantees were proposed. As a comprehensive overview in this context is not possible the following related work will focus on the ML literature. We refer the reader to a survey by \citet{ORsurvey}. 

Machine Learning has largely considered two approaches for solving routing problems. The first one autoregressively inserts one node at a time from a partial tour until a solution is complete. \citet{vinyals} proposed the Pointer-Network and used this approach to tackle the TSP. Their model was learnt from optimal examples. As these are expensive, \citet{bello, nazari} proposed approaches that used RL for optimization. Specifically, variants of REINFORCE \citep{williams} were employed and the problem catalogue was extended to the CVRP. \citet{kool} then proposed an encoder-decoder model based on the Transformer architecture and showed that good solutions can be learnt for various related problem types for instances with up to 100 nodes. \citet{falkner}  proposed an extension to incorporate time windows. While both also use variants of REINFORCE, \citet{khalil} built a graph model based on S2V optimized with Q-Learning. Instead of directly inserting the predicted next node, they choose its position with a best insertion heuristic. However their results on the TSP were only comparable with simpler constructive heuristics like farthest insertion \citep{rosenkrantz}. The autoregressive approaches have considered different search algorithms to use during inference, like Sampling, Beam Search and Active Search.

The second paradigm that was considered with ML approaches is the use of improvement heuristics. They operate on initial solutions and improve these repeatedly. Usually local search is applied. Traditionally, Meta-heuristics like Large-Neighbourhood Search (LNS) or Guided-Local Search (GLS) are used in conjunction with local search operators like 2-opt to avoid local minima. Various Meta-heuristics and local operators are implemented in \citet{ortools}. ML approaches here range from only learning the initial solution and then applying a traditional method \citep{zhao} to learning to perform local search \citep{wu, chen}. 

Other approaches that do not fit these two categories include constructing a tour non-autoregressively by predicting a heatmap of promising edges with a graph convolutional neural network \citep{yoshi}, augmenting traditional dynamic programming approaches for Routing with deep learning \citep{kool2} and changing the action space from nodes to full tours and solving the sub-problem of the best route to insert \citep{delarue}.
%
%
\section{Problem Definition}
We consider two problem variants in this paper: the CVRP and the MDVRP.
\subsubsection*{CVRP}
The problem can be described by a graph $G(N, E)$, where $N = C \cup D$ is the set of nodes consisting of customers $C = \{1,...,n\}$ and one depot $D=\{n+1\}$. Each customer $c \in C$ has a positive demand $d$. We assume a complete graph, thus the set of edges $E$ contains an edge $e_{i,j}$ representing the distance for every pair of nodes $(i, j) \in N$.  We also assume a fleet of homogeneous vehicles $K$ having the same maximum Capacity $Q$ with no restriction on the fleet size. For a valid solution all customers have to be visited exactly once. All routes must start and end at the depot. For all routes the total sum of demands must satisfy the maximum vehicle capacity. As an objective we want to minimize the total sum of distances. This formulation is also used by other works like \citep{kool, bello}. Note however that other formulations with limited, heterogeneous vehicles or other objective functions are possible.

\subsubsection*{MDVRP}
The MDVRP generalizes this problem by allowing for multiple depots $D=\{n+1, ..., n+m\}$. We require that every vehicle ends at the same depot that it has started but the model is allowed to decide at which depot a vehicle should start. Again note that other problem formulations are possible. Vehicles could be fixed and set to certain depots or allowed to return to any depot instead of the one they started for instance.

\section{Method} 
We base our architecture on the model by \citet{kool}. First we present the original model and then show which adaptions our new approach makes as well as how solutions are constructed for the CVRP and MDVRP and how the model is optimized.
\subsection{Original Attention-Model}
The attention model \citep{kool} solves routing problems using an encoder-decoder architecture and learns a policy model $\pi$ that autoregressively inserts one node at a time until a solution is complete.

\subsubsection*{Encoder}
The encoder takes the set of nodes $N$ consisting of the customer nodes and the depot node and creates an embedding of each node. The node features consist of the coordinates and the demand. These features are considered static, meaning they don't change during decoding. Thus the node embeddings are calculated once at the beginning of the episode and are reused at each step.

The encoder creates initial node embeddings $h_i$ by applying a node-wise linear projection, scaling up every node to the embedding dimension $h_{dim}=128$. $h_i$ represents the $i$-th node embedding. To differentiate between the depot node and the customer nodes, weights are not shared and two projections are learned.
\begin{equation}
h_i^{(0)}=W^{\mathrm{node}} n_i+b^{\mathrm{node}}
\end{equation}
\begin{equation}
h_0^{(0)}=W^{\mathrm{depot}} n_0+b^{\mathrm{depot}}
\end{equation}
These embeddings are then updated through $L$ attention blocks (AB):
\begin{equation}
H^{(L)}=\mathrm{AB}_L (…(\mathrm{AB}_1 (H^{(0)} )))
\end{equation}
where  $H$ is a concatenation of the initial embeddings, and $L=3$ for VRPs. Every block AB consists of a multi-head self-attention layer (MHA) \citep{vaswani}, a node-wise feed-forward layer (FF), along with batch normalization (BN) \citep{ioffe} and skip connections (skip). 
\begin{equation}
\mathrm{AB}(H^{(l)} )=\mathrm{BN}\,({\mathrm{FF^{skip}}} \,(\mathrm{BN}\,(\mathrm{MHA^{skip}} \,(H^{(l-1)} ))))
\end{equation}
where the MHA layer uses 8 heads. The $\mathrm{FF^{skip}}$ layer has one sublayer with ReLU activation and an intermediate dimension of size 512.
\begin{equation}
\mathrm{FF^{skip}}(h_i) = W_2\, \mathrm{max}(0, W_1h_i+b_1) + b_2 + h_i
\end{equation}

\subsubsection*{Decoder}
The decoder is run at every timestep $t$ and parametrizes a probability distribution over all the nodes. It takes a context $C^{(t)}=[h^{\mathrm{graph}};c^{(t)};h^{\mathrm{last}}]$ as input, where $h^{\mathrm{graph}}$ is the mean of the node embeddings, $c^{(t)}$ is the current vehicle capacity and $h^{\mathrm{last}}$ refers to the node embedding of the last position. This context is transformed with another MHA layer, where the context only attends to nodes that are feasible in this timestep. This is done with a mask $M$. Note that the context is the query and the node embeddings are the keys of this layer. In contrast the encoder computes the self attention between all pairs of nodes. The decoder instead only computes attention between context and node embeddings to arrive at a transformed context $\hat{C}^{(t)}$. This avoids the memory quadratic complexity of the MHA-layer.
\begin{equation}
\hat{C}^{(t)} = \mathrm{MHA}\,(C^{(t)}, H, M)
\label{eq:inner masking}
\end{equation}
Now, compatibility scores $u_i$ between the transformed context and the nodes are calculated as in a Single-Head Attention (SHA) mechanism where the context is again the query:
\begin{equation}
u_i = \frac{W^q\hat{C}^{(t)}W^kh_i}{\sqrt{d_k}}
\end{equation}
This gives a single value for each node and infeasible nodes are masked with $-\infty$. To arrive at the final probabilities, a softmax is applied. For more details, we refer the reader to the original paper by \citet{kool}.
\subsection{RP-DQN}
\begin{figure}[b]
	\centering
	\includegraphics[width=\linewidth, trim=0 2cm 0 2cm, clip]{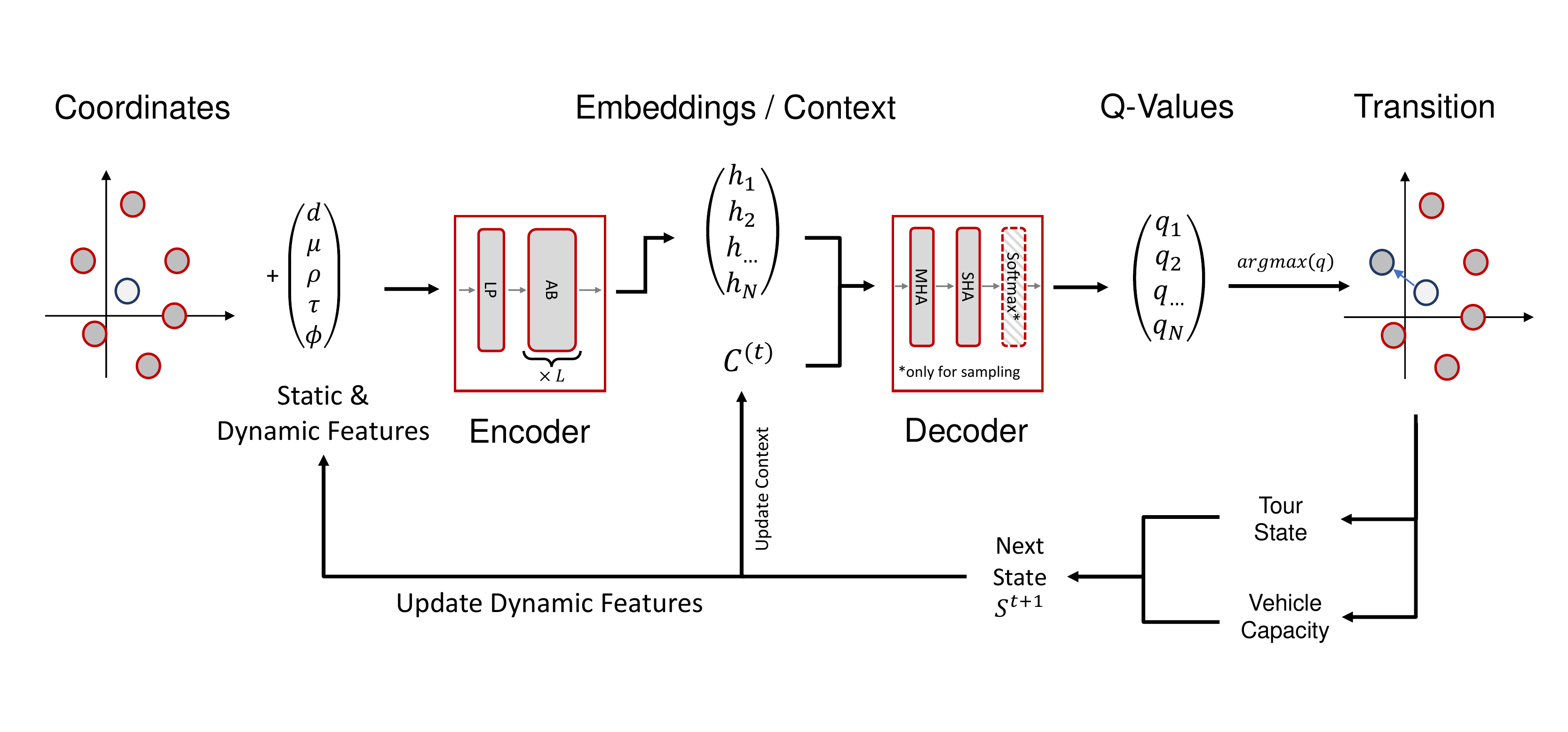}
	\caption{Architecture overview}
	\label{fig:architecture}
\end{figure}
Most of the models complexity is in the encoder, however it only considers the static components of the problem. The dynamic information that can be used in the context is quite constrained. The original model has very limited access to information about which nodes have already been visited or are infeasible due to other constraints. The only time this information is passed is in equation \ref{eq:inner masking} with an inner masking that prevents attention from infeasible nodes to the context. We feel that using dynamic node features in the encoder makes the model more expressive and utilizes its complexity better. However, in that case the encoder has to be executed in every step. As discussed already, to make this memory feasible, our model is optimized with Q-Learning. Thus, our model also has to output Q-values instead of probabilities and therefore we get rid of the softmax. The inner masking is also not needed anymore and we omit it completely.

Now, the four dynamic features $\mu_i^{(t)},\rho_i^{(t)},\tau_i^{(t)},\phi_i^{(t)}$ represent boolean variables that indicate whether at timestep $t$ the node $i$ has already been inserted in the tour, cannot currently be inserted due to capacity constraints, represents the current position of the vehicle or in the multi-depot case represents the current depot.\\
\begin{figure}
	\begin{minipage}{.5\linewidth}
		\begin{align}
		\mu_i^{(t)} &=
		\begin{cases}
		1 & \text{if } i\in \mathcal{R}^{(t)} \\
		0 &\text{otherwise} 
		\end{cases} \\[0.5cm]
		\tau_i^{(t)} &=
		\begin{cases}
		1 & \text{if } i\in \mathcal{R}^{(t)} \wedge i\notin \mathcal{R}^{(t-1)} \\
		0 &\text{otherwise} 
		\end{cases}
		\end{align}
	\end{minipage}
	\begin{minipage}{.5\linewidth}
		\begin{align}
		\rho_i^{(t)} &=
		\begin{cases}
		1 & \text{if } d_i^{(t)} > c^{(t)} \\
		0 &\text{otherwise} 
		\end{cases}\\[0.5cm]
		\phi_i^{(t)} &=
		\begin{cases}
		1 & \text{if } i \text{ is active depot} \\
		0 &\text{otherwise} 
		\end{cases}
		\end{align}
	\end{minipage}
\end{figure}
$\mathcal{R}^{(t)}$ is the set of customer nodes that have been visited at time $t$. Then the set of node features for the CVRP includes $[x_i, y_i, d_i^{(t)},\mu_i^{(t)},\rho_i^{(t)},\tau_i^{(t)}]$, where $x_i$ and $y_i$ are the coordinates of node $i$ and $d_i^{(t)}$ is the demand. $d_i^{(t)} = 0$ for nodes that have been served at timestep $t$. In the MDVRP case, $\phi_i^{(t)}$ is also added to the set of node features. The rest of the architecture stays the same. Although it potentially could be simplified, we opted to stay close to the original architecture. We show that performance improvements come mainly through our better state representation. Our model is called Routing Problem Deep Q-Network (RP-DQN). An architecture overview can be seen in Figure \ref{fig:architecture}.

\subsubsection*{MDVRP Decoding}
For the MDVRP we expand the context by the current depot embedding. This is done for our model as well as for our extension of the model by \citet{kool} to the MDVRP. Next we describe the decoding procedure. In the first step the model has not selected any node yet. We mask all customers and force it to select a depot. This depot becomes the current depot and the model selects nodes until it chooses to come back to the current depot. We don't allow for back to back depot selection and the model cannot come back to a different depot than it started from. After a route is completed the model can start another route by choosing the next depot and the procedure is repeated until all customers are inserted. The Q-values of infeasible actions are masked with $-\infty$.

\subsubsection*{Q-Learning}
To optimize our model we implemented Double DQN \citep{dqn, doubleq} with $N$-step returns and a prioritized replay buffer \citep{buffer}. We found $N=1$ to work best. For exploration, we experimented with Boltzman-Exploration \citep{sutton} with decaying softmax temperature and $\epsilon$-Greedy with decaying $\epsilon$. In the end we used $\epsilon$-Greedy and linearly decayed the rate over half of the total training episodes. Although Boltzman-Exploration initially speeds up the convergence, the model also plateaus at slightly higher costs than with $\epsilon$-Greedy.

\section{Experiments}

\subsection{Baselines}
For the CVRP we differentiate baselines from four different categories. The first category includes specialized solvers. Gurobi \citep{gurobi} was used in \citet{kool} as an optimal solver for the CVRP. This becomes intractable for instances with more than 20 customers, thus the highly optimized LKH3 \citep{lkh3} was employed by \citet{kool} alternatively. It transforms constrained problems into TSPs via penalty functions and other means and then solves those heuristically. The second category consist of ML approaches that construct solutions sequentially and use greedy inference. We compare our model with the approaches by \citet{kool, nazari, falkner}. The third category also includes ML based approaches with sequential construction, but this time using sampling or beam search inference. We compare with the same models as in the previous category. Our model, \citet{kool} and \citet{falkner} use sampling while \citet{nazari} uses beam search. The last category encompasses improvement heuristics that use local search. We include it for completeness sake. They operate on an initial solution and improve it iteratively by selecting solutions in the neighborhood. OR-Tools \citep{ortools} is a classical approach using meta-heuristics while \citet{wu} is a machine learning approach. For the MDVRP, no ML baselines are available. Thus, we adapt the model by \citet{kool} to support MDVRP and train it ourselves. We set up the decoding procedure and context as for our model. For a traditional baseline we stick to Google OR-Tools as it provides a flexible modeling framework. However OR-Tool does not provide an out of the box MDVRP solver, thus we use the CVRP framework and simulate the MDVRP by setting different starting and finishing points to the vehicles. 
\subsection{Data}
For the training of the model we will generate data. We will create data with $|C| = {20,50,100}$ and train a model for each of these. In the main experiments, the problem size stays the same for training and testing. Additionally, we conducted a generalization study that shows how the models perform on different problem sizes, which can be seen in Appendix \ref{appendix:generalization}. Since the data can be generated unlimitedly, we will use every problem instance only once and generate new data after every episode. Note that the data can be generated with various different properties. For instance, the customers can have a special alignment in grids or star-like structures, be sampled uniformly or according to some other distribution. In order to compare, our data generation follows \citet{nazari}. Their dataset is used by most of the ML literature. For testing we have used the exact test set that was provided by \citet{nazari} for each problem size. It consists of 10,000 instances that were generated with the same properties as the training data. It uses the euclidean distance. Demands $d \in [1,9]$ are sampled uniformly. Coordinates are sampled uniformly from the unit square. Vehicles have capacity 30, 40 and 50 depending on the problem size. For the MDVRP we generate the data in the same fashion and create our own test set.
\subsection{CVRP Results}
Table \ref{table:CVRP} shows that our approach outperforms all other models that construct the solution by sequentially selecting nodes. This applies for both greedy and sampling inference. Only the improvement method by \citet{wu} achieves better results. However, improvement heuristics operate on an initial solution and will try to improve this repeatedly. It has been shown that these approaches scale to better final performance when given a better initial starting solution \citep{zhao}. Thus better approaches to find initial solutions like ours can be used to kick-start improvement methods like OR-Tools or the machine learning approach by \citet{wu}. We also notice that our percentage gain increases with the problem size, stressing the importance of dynamic features for larger, real world problems. Appendix \ref{appendix:time} gives a more detailed comparison including the timings of some of the methods.\\
\newcommand{\myrowwidth}{8em}
\begin{table}[h]
	\small
	\centering
	\caption{CVRP Results}
	\begin{tabular}{cccccccc}
		\Xhline{2\arrayrulewidth}
		\multicolumn{2}{c|}{\multirow{6}{*}[1.6em]{Method}} & \multicolumn{6}{c}{\textbf{Problem Size}}  	\\ \cline{3-8}
		\multicolumn{2}{c|}{}&\multicolumn{2}{c}{\textbf{20}}&\multicolumn{2}{c}{\textbf{50}}&\multicolumn{2}{c}{\textbf{100}}													\\
		\multicolumn{2}{c|}{}&Mean&Gap \%&Mean&Gap \%&Mean&Gap \%												\\ \hline \hline
		
		\multicolumn{1}{l}{\multirow{2}{\myrowwidth}[-2pt]{Specialized Solver}}&	\multicolumn{1}{l|}{Gurobi}						&6.1	&0.00	&-	&-	&-	&-				\\
		\multicolumn{1}{l}{}&												\multicolumn{1}{l|}{LKH3}								&6.14	&0.66	&10.38	&0.00	&15.65	&0.00		\\ 
		\hline
		
		\multicolumn{1}{l}{\multirow{3}{\myrowwidth}[-14pt]{Sequential Policy (Greedy)}}&	\multicolumn{1}{l|}{RP-DQN}	&\textbf{6.36}	&4.26	&\textbf{10.92}	&5.20	&\textbf{16.59}	&6.01				\\
		\multicolumn{1}{l}{}&																	\multicolumn{1}{l|}{Kool}			&6.4	&4.92	&10.98	&5.78	&16.8	&7.35		\\
		\multicolumn{1}{l}{}&																	\multicolumn{1}{l|}{Falkner}			&6.47	&6.07	&11.44	&10.21	&-	&-		\\ 
		\multicolumn{1}{l}{}&																	\multicolumn{1}{l|}{Nazari}			&6.59	&8.03	&11.39	&9.73	&17.23	&10.10		\\ 
		\hline
		
		\multicolumn{1}{l}{\multirow{3}{\myrowwidth}[-2pt]{Sequential Policy (Sampling/Beam search)}}&	\multicolumn{1}{l|}{RP-DQN (1024s)}	&\textbf{6.24}	&2.30	&\textbf{10.59}	&2.02	&\textbf{16.11}	&2.94				\\
		\multicolumn{1}{l}{}&															\multicolumn{1}{l|}{Kool (1280s)}			&6.25	&2.46	&10.62	&2.31	&16.23	&3.71		\\
		\multicolumn{1}{l}{}&															\multicolumn{1}{l|}{Falkner (1280s)}			&6.26	&2.62	&10.84	&4.43	&-	&-		\\
		\multicolumn{1}{l}{}&															\multicolumn{1}{l|}{Nazari (10bs)}			&6.4	&5.41	&11.31	&8.96	&17.16	&9.65		\\  
		\hline
		
		\multicolumn{1}{l}{\multirow{2}{\myrowwidth}[-1pt]{Local Search}}&	\multicolumn{1}{l|}{Wu (5000 steps)}							&\textbf{6.12}	&0.33	&\textbf{10.45}	&0.67	&\textbf{16.03}	&2.43				\\
		\multicolumn{1}{l}{}&											\multicolumn{1}{l|}{OR-Tools}								&6.43	&5.41	&11.31	&8.96	&17.16	&9.65		\\ 
		\Xhline{2\arrayrulewidth}
	\end{tabular}
	\label{table:CVRP}
\end{table}
\subsection{MDVRP Results}
In Table \ref{table:MDVRP} an even greater lift as for the CVRP can be seen. The model by \citet{kool} reaches only subpar performance when the problem size increases. We attribute the results to the more powerful state representation of our model. We assume that this has more impact on the MDVRP as it is a more complicated problem than the CVRP. We also want to highlight that training our model is much more memory efficient compared to the standard model by \citet{kool}.
\begin{table}[h]
	\small
	\centering
	\caption{MDVRP Results}
	\begin{tabular}{c|cccccc}
		\Xhline{2\arrayrulewidth}
		\multirow{3}{*}[-5pt]{Method} & \multicolumn{6}{c}{\textbf{Problem Size}}  	\\ \cline{2-7}
		&\multicolumn{2}{c}{\textbf{20}}&\multicolumn{2}{c}{\textbf{50}}&\multicolumn{2}{c}{\textbf{100}}													\\
		&Mean&Gap \%&Mean&Gap \%&Mean&Gap \%												\\ \hline \hline
		RP-DQN - Greedy&5.48&2.62&8.04&4.15&11.99&4.08\\
		Kool - Greedy&5.68&6.37&8.84&14.51&13.17&14.32\\
		RP-DQN - Sampling 1024&\textbf{5.34}&0.00&\textbf{7.72}&0.00&\textbf{11.52}&0.00\\
		Kool - Sampling 1024&5.42&1.50&8.11&5.05&12.15&5.47\\
		OR-Tools&6.74&26.23&9.02&16.84&12.92&12.15\\ 
		\Xhline{2\arrayrulewidth}	
	\end{tabular}
	\label{table:MDVRP}
\end{table}
\subsection{Learning Curves}
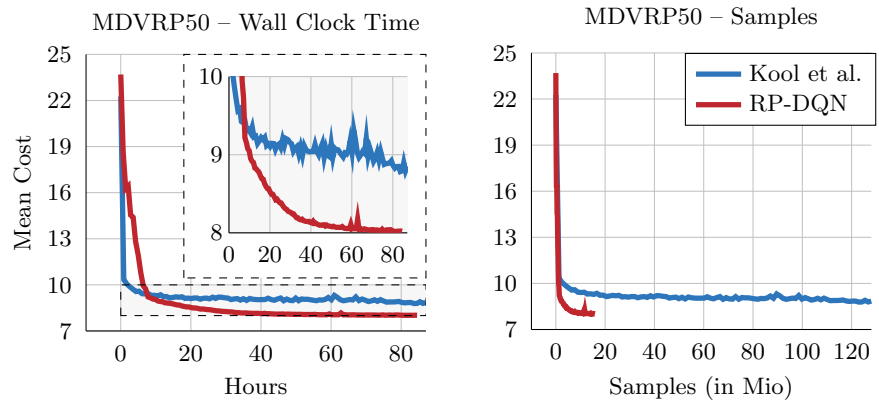
\begin{figure}[h]
	\definecolor{myblue1}{cmyk}{0.80,0.50,0.0,0.0}
	\definecolor{myblue2}{cmyk}{0.56,0.35,0.0,0.0}
	\definecolor{myblue3}{cmyk}{0.40,0.25,0.0,0.0}
	\definecolor{myblue4}{cmyk}{0.24,0.15,0.0,0.0}
	\definecolor{myblue5}{cmyk}{0.12,0.075,0.0,0.0}
	
	\definecolor{myred1}{cmyk}{0.25,1.0,1.0,0.0}
	\definecolor{myred2}{cmyk}{0.175,0.70,0.70,0.0}
	\definecolor{myred3}{cmyk}{0.125,0.50,0.50,0.0}
	\definecolor{myred4}{cmyk}{0.075,0.30,0.30,0.0}
	\definecolor{myred5}{cmyk}{0.0375,0.15,0.15,0.0}
	\newcommand{\boxxleft}{1.9cm}
	\newcommand{\boxxtop}{1.78cm}
	\newcommand{\boxCL}{0.6cm}
	\begin{minipage}{.5\textwidth}
		\centering
		\begin{tikzpicture}
		\begin{axis}[
		at={(0,0)},
		tick style={draw=none},
		title={MDVRP50 -- Wall Clock Time},
		xlabel={Hours},
		ylabel={Mean Cost},
		width={\linewidth},
		grid = major,
		major grid style={line width=.2pt,draw=gray!50},
		axis lines=left,
		axis line style={-},
		xmin=-9.9,
		ymin=7,
		ymax=25,
		every axis plot/.append style={line width=2pt},
		xtick = {0,20,...,100},
		ytick={7,10,13,...,25},
		]
		\begin{scope}[on background layer] 
		\draw [dashed,fill=gray!6,opacity=1, draw=none] (axis cs:0,8) rectangle (axis cs:87.1,10);
		\end{scope}
		\addplot[color=myblue1] table [x=hours ,y=Value, col sep=comma, each nth point=1, mark=none] {kool.csv};
		\addplot[color=myred1] table [x=hours ,y=Value, col sep=comma, each nth point=1, mark=none] {ours.csv};
		\draw [dashed] (axis cs:0,8) rectangle (axis cs:87.1,10);
		\end{axis}

		\begin{axis}[
		no markers,
		at={(\boxxleft,\boxxtop-13.5)},
		axis background/.style={fill=gray!6},
		tick style={draw=none},
		grid = major,
		major grid style={line width=.2pt,draw=gray!50},
		axis lines=left,
		axis line style={-},
		xmin=0,
		ymin=8,
		ymax=10,
		height={0.55\linewidth},
		width={\linewidth - \boxxleft - \boxCL},
		every axis plot/.append style={line width=2pt},
		xtick = {0,20,...,100},
		xticklabels=,
		ytick={7,8,9,10,13,...,25},
		]
		\addplot[color=myblue1] table [x=hours ,y=Value, col sep=comma, each nth point=1, mark=none] {kool.csv};
		\addplot[color=myred1] table [x=hours ,y=Value, col sep=comma, each nth point=1, mark=none] {ours.csv};
		\end{axis}
		
		\draw [fill=white, dashed]let \p1 = (current axis.south west), \p2 = (current axis.north east) in (\x1-\boxCL,\y1-\boxCL)rectangle (\x2+\boxCL, \y2+\boxCL);
		\begin{axis}[
		no markers,
		label style={font=\small, align=right},
		y tick label style={at={(yticklabel cs:-0.05,0)},
			anchor=near yticklabel,
			inner sep=0pt,},
		ticklabel style = {font=\small},
		at={(\boxxleft,\boxxtop-13.5)},
		axis background/.style={fill=gray!6},
		tick style={draw=none},
		grid = major,
		major grid style={line width=.2pt,draw=gray!50},
		axis lines=left,
		axis line style={-},
		xmin=0,
		ymin=8,
		ymax=10,
		height={0.6\linewidth},
		width={\linewidth - \boxxleft - \boxCL +10},
		every axis plot/.append style={line width=2pt},
		xtick = {0,20,...,100},
		ytick={7,8,9,10,13,...,25},
		]
		\addplot[color=myblue1] table [x=hours ,y=Value, col sep=comma, each nth point=1, mark=none] {kool.csv};
		\addplot[color=myred1] table [x=hours ,y=Value, col sep=comma, each nth point=1, mark=none] {ours.csv};
		\end{axis}
		\end{tikzpicture}
		
	\end{minipage}
	\begin{minipage}{.5\textwidth}
		\centering
		\begin{tikzpicture}
		\begin{axis}[
		no markers,
		at={(0,0)},
		tick style={draw=none},
		title={MDVRP50 -- Samples},
		xlabel={Samples (in Mio)},
		width={\linewidth},
		grid = major,
		major grid style={line width=.2pt,draw=gray!50},
		axis lines=left,
		axis line style={-},
		xmin=-9.9,
		ymin=7,
		ymax=25,
		every axis plot/.append style={line width=2pt},
		xtick = {0,20,...,130},
		ytick={7,10,13,...,25},
		legend entries={Kool et al., RP-DQN},
		legend cell align=left,
		legend style = {at={(axis cs:130,25)},anchor=north east},
		]
		
		\addplot[color=myblue1] table [x expr=\thisrow{samples}*1e-6 ,y=Value, col sep=comma, each nth point=1, mark=none] {kool.csv};
		\addplot[color=myred1] table [x expr=\thisrow{samples}*1e-6 ,y=Value, col sep=comma, each nth point=1, mark=none] {ours.csv};
		\end{axis}
		\end{tikzpicture}
	\end{minipage}
	\caption{Learning Curves for the MDVRP with 50 customers comparing the wall clock time and sample efficiency.}
	\label{fig:learning curves}
\end{figure}

In Figure \ref{fig:learning curves} we exemplify the learning behavior of both models on the MDVRP with 50 customers run on the same hardware. Due to the use of a buffer, the sample efficiency is improved greatly. While this is less important for simple environments, it has significant impact for problems that are more expensive to compute. For routing, this can include environments with more hard and soft constraints, stochastic components and objective functions that trade-off multiple goals. Figure \ref{fig:learning curves} shows that the learning of our model starts off slower, however we always converge significantly faster.
\section{Conclusion}
In this paper we present a new approach to tackle complex routing problems based on learning from temporal differences, specifically through Q-Learning, when optimizing autoregressive policies that sequentially construct solutions by inserting one node at a time for solving routing problems. We showed that this learning procedure allows the incorporation of dynamic node features, enabling more powerful models which lead to state-of-the-art performance on the CVRP for autoregressive policies. Additionally, the sample efficiency is greatly improved. Although our model still falls short of specialized solvers like LKH3 and improvement methods, it is useful to find very good initial solutions. Future work could include combining powerful initial solution finders like RP-DQN with improvement heuristics.
We also demonstrated that the dynamic components become more important for the MDVRP, a problem type that was not explored with RL before. We assume that this holds for other more complicated problem types like CVRP with time windows. Future work could include extending our model to these problem types. Further, more work is needed to improve generalizability. This could include training on broader distributions and building automatic data generators from real life datasets.
\subsubsection{Acknowledgement}
This work is co-funded via the research project L2O\footnote{\url{https://www.ismll.uni-hildesheim.de/projekte/l2o_en.html}} funded by the German Federal Ministry of Education and Research (BMBF) under the grant agreement no. 01IS20013A and the European Regional Development Fund project TrAmP\footnote{\url{https://www.ismll.uni-hildesheim.de/projekte/tramp.html}} under the grant agreement no. 85023841.
\appendix
\section{Runtime Comparison}\label{appendix:time}
Reporting the runtime is difficult as it can differ in order of magnitudes due to implementation differences (C++ vs Python) and hardware. \citet{kool} thus decided to report the runtime over the complete test set of 10,000 instances. Other literature like \citet{wu} has followed them. We feel that this adds another layer of obscurity over the runtime comparison as they then decided to parallelize over multiple instances. ML approaches were parallelized via batch computation. Most traditional methods are single threaded CPU applications. They were parallelized over instances by launching more threads if the CPU has them available. We feel that for practical applications if enough instances have to be solved for instance parallelization to be useful, the corresponding hardware can be bought. More important is the time it takes to solve a single instance or alternatively the time it takes to solve the 10,000 instances when not parallelizing over instances. In Table \ref{table:CVRP timings} we collect some timings reported by the literature and our own. Note that the timings are not directly comparable due to the discussed reasons. Also note that most methods have some way of trading performance and time off. Beam Searches can have bigger width, more solutions can be sampled and improvement methods can make more steps. Additionally, none of the methods will improve indefinitely but hit diminishing returns instead. Thus there are more and less reasonable spots to trade-off. Ultimately however, this is also application dependent. Reporting full trade-off curves for all of the methods is not possible.
\begin{table}[h]
	\small
	\centering
	\begin{threeparttable}
		\caption{CVRP Results and Timings to solve all 10,000 instances}
		\begin{tabular}{l|cccccc}
			\Xhline{2\arrayrulewidth}
			\multirow{3}{*}[-4pt]{Method} & \multicolumn{6}{c}{\textbf{Problem Size}}	\\ \cline{2-7}
			&\multicolumn{2}{c}{\textbf{20}}&\multicolumn{2}{c}{\textbf{50}}&\multicolumn{2}{c}{\textbf{100}}	\\
			&Mean&Time& Mean&Time& Mean&Time	\\ \hline \hline
			Gurobi			&6.1	&-	&-	&-	&-	&-						\\
			LKH3			&6.14	&(2h$\ssymbol{1}$)	&10.38	&(7h$\ssymbol{1}$)	&15.65	&(13h$\ssymbol{1}$)		\\ \hline
			RP-DQN -- Greedy			&6.36	&(3s$\ssymbol{2}$/10min$\ssymbol{8}$)		&10.92	&(14s$\ssymbol{2}$/30min$\ssymbol{8}$)	&16.59	&(50s$\ssymbol{2}$/78min$\ssymbol{8}$)		\\
			Kool -- Greedy			&6.4	&(1s$\ssymbol{2}$)	&10.98	&(3s$\ssymbol{2}$)	&16.8	&(8s$\ssymbol{2}$)		\\
			RP-DQN -- 1024s	&6.24	&(52m$\ssymbol{7}$)		&10.59	&(5h$\ssymbol{7}$)	&16.11	&(15h$\ssymbol{7}$)		\\
			Kool -- 1280s	&6.25	&(6m$\ssymbol{7}$)	&10.62	&(28m$\ssymbol{7}$)	&16.23	&(2h$\ssymbol{7}$)		\\ \hline
			Wu (5000 steps) &6.12	&(2h$\ssymbol{2}$)		&10.45	&(4h$\ssymbol{2}$)	&16.03	&(5h$\ssymbol{2}$)		\\
			OR-Tools		&6.43	&(2m$\ssymbol{3}$)		&11.31	&(13m$\ssymbol{3}$)	&17.16	&(46m$\ssymbol{3}$)		\\ 
			\Xhline{2\arrayrulewidth}
		\end{tabular}
		\label{table:CVRP timings}
		\begin{tablenotes}
			\small
			\item $\ssymbol{1}$32 instances were solved in parallel on two CPUs \citep{kool}.
			\item $\ssymbol{2}$Time for solving many instances in parallel through GPU batch computation.
			\item $\ssymbol{3}$Time reported by \citet{wu}. Only one Thread.
			\item $\ssymbol{7}$Time for solving one instance at a time.
			\item $\ssymbol{8}$Time for solving one instance at a time on CPU
		\end{tablenotes}
	\end{threeparttable}
\end{table}\\
Our method is extremely quick in greedy and then takes more time the more samples are used. This behaviour is respectively the same for the Kool model although our model expectedly takes more time as the encoder is run at every step. 15 hours for the size 100 model with sampling seems high, however consider that one instance was solved at a time, thus a single instance still only takes less than 6 seconds. The only two methods that have better results are LKH3 and \citet{wu}'s improvement approach. For both the time was only reported with a high degree of instance parallelization, thus it should be expected that our model takes less time than them on a fair single instance comparison.

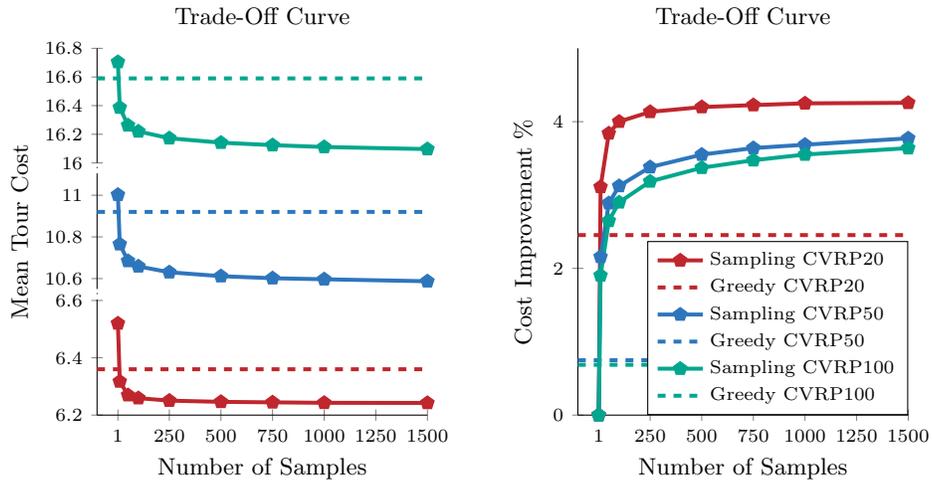
\begin{figure}
	\pgfplotsset{every tick label/.append style={font=\scriptsize}}
	\pgfplotsset{every axis label/.append style={font=\small}}
	\pgfplotsset{every axis title/.append style={font=\small}}
	\begin{tikzpicture}
	\newcommand{\xmax}{1500}
	\newcommand{\xmin}{-100}
	\definecolor{mygreen1}{cmyk}{1.0, 0.0, 0.6, 0.0}
	\definecolor{myblue1}{cmyk}{0.80,0.50,0.0,0.0}
	\definecolor{myred1}{cmyk}{0.25,1.0,1.0,0.0}
	\begin{axis}[
	/pgf/number format/1000 sep={},
	name=a,
	scale only axis,
	xlabel={Number of Samples},
	width=.36\linewidth,
	height=0.125\linewidth,
	axis x line=bottom,
	axis y line=box,
	xmin=\xmin, xmax=\xmax,
	ymin=6.2, ymax=6.6,
	xtick = {1,250,500, ..., 1500},
	every axis plot/.append style={line width=1.5pt},
	ytick = {6.2,6.4,6.6},
	axis lines=left,
	axis line style={-},
	]
	
	\addplot[color=myred1, mark=pentagon*, mark options={scale=1,solid},] 			table [x=samples, y=mean, col sep=semicolon, each nth point=1,] {tradeoff20.csv};
	\addplot[color=myred1, mark=none, dashed, domain=-100:1600, samples=2] {6.36};
	\end{axis}
	\begin{axis}[
	name=b,
	scale only axis,
	ylabel={Mean Tour Cost},
	width=.36\linewidth,
	height=0.125\linewidth,
	yshift=0.1375\linewidth,
	axis x line=none,
	axis y line=box,
	xmin=\xmin, xmax=\xmax,
	ymin=10.55, ymax=11.1,
	every axis plot/.append style={line width=1.5pt},
	axis lines=left,
	axis line style={-},
	]
	
	\addplot[color=myblue1, mark=pentagon*, mark options={scale=1,solid},] 			table [x=samples, y=mean, col sep=semicolon, each nth point=1,] {tradeoff50.csv};
	\addplot[color=myblue1, mark=none, dashed, domain=-100:1600, samples=2] {10.92};
	\end{axis}
	\begin{axis}[
	name=c,
	scale only axis,
	title={Trade-Off Curve},
	width=.36\linewidth,
	height=0.125\linewidth,
	yshift=0.275\linewidth,
	axis x line=none,
	axis y line*=box,
	xmin=\xmin, xmax=\xmax,
	ymin=16, ymax=16.8,
	xticklabel=\empty,
	every axis plot/.append style={line width=1.5pt},
	axis lines=left,
	axis line style={-},
	]
	\addplot[color=mygreen1, mark=pentagon*, mark options={scale=1,solid},] 			table [x=samples, y=mean, col sep=semicolon, each nth point=1,] {tradeoff100.csv};
	\addplot[color=mygreen1, mark=none, dashed, domain=-100:1600, samples=2] {16.59};
	\end{axis}
	
	\draw [dotted] (a.north west) -- (b.south west);
	\draw [dotted] (b.north west) -- (c.south west);
	
	\begin{axis}[
	/pgf/number format/1000 sep={},
	align = center,
	scale only axis,
	at={($(a.south east)+(2cm,0)$)},
	anchor = south west,
	title={Trade-Off Curve},
	xlabel={Number of Samples},
	ylabel={Cost Improvement \%},
	width={0.36\linewidth},
	height={0.4 \linewidth},
	axis lines=left,
	axis line style={-},
	xmin=-100,
	xmax=1500,
	ymin=0,
	ymax=5,
	every axis plot/.append style={line width=1.5pt},
	xtick = {1,250,500, ..., 1500},
	legend entries={Sampling CVRP20, Greedy CVRP20, Sampling CVRP50, Greedy CVRP50, Sampling CVRP100, Greedy CVRP100},
	legend cell align=left,
	legend style = {at={(axis cs:1500,0)},anchor=south east,font=\scriptsize},
	]
	\addplot[color=myred1, mark=pentagon*, mark options={scale=1,solid},] 			table [x=samples, y=normalized, col sep=semicolon, each nth point=1,] {tradeoff20.csv};
	\addplot[color=myred1, mark=none, dashed, domain=-100:1600, samples=2] {2.453823368};
	
	\addplot[color=myblue1, mark=pentagon*, mark options={scale=1,solid},] 			table [x=samples, y=normalized, col sep=semicolon, each nth point=1,] {tradeoff50.csv};
	\addplot[color=myblue1, mark=none, dashed, domain=-100:1600, samples=2] {0.748010618};
	
	\addplot[color=mygreen1, mark=pentagon*, mark options={scale=1,solid},] 			table [x=samples, y=normalized, col sep=semicolon, each nth point=1,] {tradeoff100.csv};
	\addplot[color=mygreen1, mark=none, dashed, domain=-100:1600, samples=2] {0.685569965};
	
	\end{axis}
	\end{tikzpicture}
	\caption{This figure shows how much additional samples improve the solution quality.}
	\label{fig:tradeoff}
\end{figure}

Additionally, 1024 samples is already at a point on the trade-off curves where there are negligible performance improvements as can be seen in figure \ref{fig:tradeoff}.

\section{Generalization Study}\label{appendix:generalization}
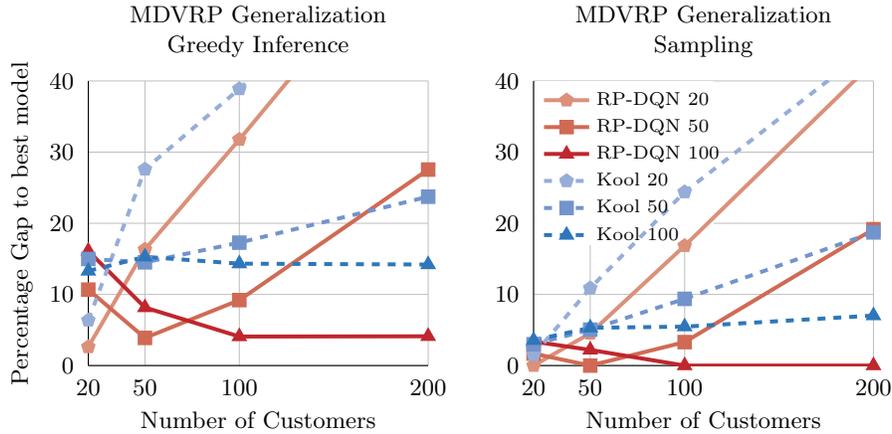
\begin{figure}[]
	\definecolor{myblue1}{cmyk}{0.80,0.50,0.0,0.0}
	\definecolor{myblue2}{cmyk}{0.56,0.35,0.0,0.0}
	\definecolor{myblue3}{cmyk}{0.40,0.25,0.0,0.0}
	\definecolor{myblue4}{cmyk}{0.24,0.15,0.0,0.0}
	\definecolor{myblue5}{cmyk}{0.12,0.075,0.0,0.0}
	
	\definecolor{myred1}{cmyk}{0.25,1.0,1.0,0.0}
	\definecolor{myred2}{cmyk}{0.175,0.70,0.70,0.0}
	\definecolor{myred3}{cmyk}{0.125,0.50,0.50,0.0}
	\definecolor{myred4}{cmyk}{0.075,0.30,0.30,0.0}
	\definecolor{myred5}{cmyk}{0.0375,0.15,0.15,0.0}
	\newcommand{\genheight}{0.88\linewidth}
	
	\begin{minipage}{.5\linewidth}
		\centering
		\begin{tikzpicture}
		\begin{axis}[
		align = center,
		at={(0,0)},
		tick style={draw=none},
		title={MDVRP Generalization \\ Greedy Inference},
		xlabel={Number of Customers},
		ylabel={Percentage Gap to best model},
		width={\linewidth},
		height={\genheight},
		grid = major,
		major grid style={line width=.2pt,draw=gray!50},
		axis lines=left,
		axis line style={-},
		ymin=0,
		ymax=40,
		every axis plot/.append style={line width=1.5pt},
		xtick = {20,50,100,200},
		]
		
		\addplot[color=myred3, mark=pentagon*, mark options={scale=1,solid},] 			table [x=Dataset Size, y=Ours20 g, col sep=semicolon, each nth point=1,] {generalization_mdvrp.csv};
		\addplot[color=myred2, mark=square*, mark options={scale=1,solid},] 			table [x=Dataset Size, y=Ours50 g, col sep=semicolon, each nth point=1] {generalization_mdvrp.csv};
		\addplot[color=myred1, mark=triangle*, mark options={scale=1,solid},]  			table [x=Dataset Size, y=Ours100 g, col sep=semicolon, each nth point=1] {generalization_mdvrp.csv};
		\addplot[color=myblue3, dashed, mark=pentagon*, mark options={scale=1,solid},] 	table [x=Dataset Size, y=kool20 g, col sep=semicolon, each nth point=1,] {generalization_mdvrp.csv};
		\addplot[color=myblue2, dashed, mark=square*, mark options={scale=1,solid},] 	table [x=Dataset Size, y=kool50 g, col sep=semicolon, each nth point=1,] {generalization_mdvrp.csv};
		\addplot[color=myblue1, dashed, mark=triangle*, mark options={scale=1,solid},] 	table [x=Dataset Size, y=kool100 g, col sep=semicolon, each nth point=1, ] {generalization_mdvrp.csv};
		\end{axis}
		\end{tikzpicture}
	\end{minipage}
	\begin{minipage}{.5\linewidth}
		\centering
		\begin{tikzpicture}
		\begin{axis}[
		align = center,
		at={(0,0)},
		tick style={draw=none},
		title={MDVRP Generalization \\ Sampling},
		xlabel={Number of Customers},
		width={\linewidth},
		height={\genheight},
		grid = major,
		major grid style={line width=.2pt,draw=gray!50},
		axis lines=left,
		axis line style={-},
		ymin=0,
		ymax=40,
		every axis plot/.append style={line width=1.5pt},
		xtick = {20,50,100,200},
		legend entries={RP-DQN 20, RP-DQN 50, RP-DQN 100, Kool 20, Kool 50, Kool 100},
		legend cell align=left,
		legend style = {at={(axis cs:20,40)},anchor=north west,font=\scriptsize, draw=none, fill=none},
		]
		\draw [fill=white, draw=none] (axis cs:20.43,39.94) rectangle (axis cs:115, 17);
		\addplot[color=myred3, mark=pentagon*, mark options={scale=1,solid},] 			table [x=Dataset Size, y=Ours20 s, col sep=semicolon, each nth point=1,] {generalization_mdvrp.csv};
		\addplot[color=myred2, mark=square*, mark options={scale=1,solid},] 			table [x=Dataset Size, y=Ours50 s, col sep=semicolon, each nth point=1] {generalization_mdvrp.csv};
		\addplot[color=myred1, mark=triangle*, mark options={scale=1,solid},]  			table [x=Dataset Size, y=Ours100 s, col sep=semicolon, each nth point=1] {generalization_mdvrp.csv};
		\addplot[color=myblue3, dashed, mark=pentagon*, mark options={scale=1,solid},] 	table [x=Dataset Size, y=kool20 s, col sep=semicolon, each nth point=1,] {generalization_mdvrp.csv};
		\addplot[color=myblue2, dashed, mark=square*, mark options={scale=1,solid},] 	table [x=Dataset Size, y=kool50 s, col sep=semicolon, each nth point=1,] {generalization_mdvrp.csv};
		\addplot[color=myblue1, dashed, mark=triangle*, mark options={scale=1,solid},] 	table [x=Dataset Size, y=kool100 s, col sep=semicolon, each nth point=1, ] {generalization_mdvrp.csv};
		\end{axis}
		\end{tikzpicture}
	\end{minipage}
	\caption{This figure shows the generalization to different problem sizes.}
	\label{fig:mdvrp generalization}
\end{figure}

This study tests the ability to generalize. All models were trained on one size and tested on instances between 20 and 200 nodes. Figure \ref{fig:mdvrp generalization} shows the percentage gap to the best model from both inference settings.

\end{document}